\def\year{2019}\relax
\documentclass[letterpaper]{article} 
\usepackage{aaai19}  
\usepackage{times}  
\usepackage{helvet}  
\usepackage{courier}  
\usepackage{url}  
\usepackage{graphicx}  
\usepackage{amsmath}
\usepackage[vlined, ruled, boxed]{algorithm2e}
\usepackage{multirow}
\usepackage{booktabs}

\pdfoutput=1

\begin{document}

\title{Deep Optimisation:\\ Solving Combinatorial Optimisation Problems using Deep Neural Networks}

\author{J.R. Caldwell, R.A Watson \ and C. Thies\\
Agents, Interaction and Complexity\\
University of Southampton\\
SO17 1BJ, U.K.
\And
J.D. Knowles\\
School of Computer Science\\
University of Birmingham\\
B15 2TT, U.K.
}

\maketitle
\begin{abstract}
Deep Optimisation (DO) combines evolutionary search with Deep Neural Networks (DNNs) in a novel way - not for optimising a learning algorithm, but for finding a solution to an optimisation problem. Deep learning has been successfully applied to classification, regression, decision and generative tasks and in this paper we extend its application to solving optimisation problems. Model Building Optimisation Algorithms (MBOAs), a branch of evolutionary algorithms, have been successful in combining machine learning methods and evolutionary search but, until now, they have not utilised DNNs. DO is the first algorithm to use a DNN to learn and exploit the problem structure to adapt the variation operator (changing the neighbourhood structure of the search process). We demonstrate the performance of DO using two theoretical optimisation problems within the MAXSAT class. The Hierarchical Transformation Optimisation Problem (HTOP) has controllable deep structure that provides a clear evaluation of how DO works and why using a layerwise technique is essential for learning and exploiting problem structure. The Parity Modular Constraint Problem (MC\textsubscript{parity}) is a simplistic example of a problem containing higher-order dependencies (greater than pairwise) which DO can solve and state of the art MBOAs cannot. Further, we show that DO can exploit deep structure in TSP instances. Together these results show that there exists problems that DO can find and exploit deep problem structure that other algorithms cannot. Making this connection between DNNs and optimisation allows for the utilisation of advanced tools applicable to DNNs that current MBOAs are unable to use.

\end{abstract}

\section{Introduction}
Combinatorial optimisation (CO) is the task of searching for a solution from a finite collection of possible candidate solutions that maximises the objective function. Put differently, the task is to reduce the large finite collection of possible solutions to a single (or small number of) optimal solution(s). In some cases, CO problems require methods that either have a bias to the problem structure or can learn the problem structure during the optimisation process such that it can be exploited. This hidden problem structure is caused by variable correlations and variable decompositions (building-blocks/modules \cite{goldberg1989genetic,holland1992adaptation}) and is, generally, unknown. The hidden structure can contain a multitude of characteristics such as near-separable decomposition, hierarchy, overlapping linkage and, as this paper shows, deep structure.

Deep learning (DL) is tasked with learning high-order representations (features) of a data-set that construct an output to satisfy the learning objective. The higher-order features are constructed from a sub-set of units from the layer below. DL performs this recursively, reducing the dimensionality of the visible space and generating an organised hierarchical structure. Deep Neural Networks (DNNs) are capable of learning complex high-order features from unlabelled data. Evolutionary search has been used in conjunction with DNNs, namely to decide on network topological features (number of hidden layers, nodes per a layer etc) \cite{stanley2002evolving} and for evolving weights of a DNN \cite{such2017deep} (neuro-evolution). However, DO is different; whereas previous methods use optimisers to improve the performance of a learning algorithm, DO is the reverse - it uses learning to improve the performance of an optimisation algorithm (`learning how to optimise, not optimising how to learn').

In CO it is a common intuition that solutions to small sub-problems can be combined together to solve larger sub-problems, and that this can proceed through multiple levels until the whole problem is solved. However, in practice, this is difficult to achieve (without expert domain knowledge) because the problem structure necessary for such problem decomposition is generally unknown. In learning, it is a common intuition that concepts can be learned by combining low-level features to discover higher-level features, and that this can proceed through multiple levels until the high-level concepts are found. DO brings these two together so that multi-level learning can discover multi-level problem structure automatically.

Model-Building Optimisation Algorithms (MBOAs), also known as Estimation of Distribution Algorithm's (EDAs) \cite{hauschild2011introduction} are black-box solvers, inspired by biological evolutionary processes, that solve CO problems by using machine learning techniques to learn and exploit the hidden problem structure. MBOAs work by learning correlations present in a sample of fit candidate solutions and construct a model that captures the multi-variate correlations which, if learnt successfully, represents the hidden problem structure. They then proceed to generate new candidate solutions by exploiting this learnt information enabling them to find solutions that are otherwise pathologically difficult to find. It is the ability of MBOAs to exploit the hidden structure that has brought their success to solving optimisation problems \cite{aickelin2007estimation,santana2008protein,pelikan2005hierarchical,goldman2014parameter,thierens2010linkage}. Models used in MBOAs include Bayesian Networks \cite{pelikan1999boa,pelikan2005hierarchical}, Dependency Matrices \cite{hsu2015optimization}, and Linkage Trees \cite{thierens2010linkage,goldman2014parameter}.

The model is used for two tasks in MBOAs: 1) To learn, unsupervised, correlations between variables as to form higher-orders of organisation reducing the dimensionality of the solution space from combinations of individual variables to combinations of module solutions (features). 2) To generate or modify new candidate solutions in a way that exploits this learnt information. Thus far, the models used in MBOAs are simplistic in comparison to the state of the art models used by the machine learning community. We believe DNNs contain all the necessary characteristics required for solving CO problems and fit naturally as the model role in MBOAs.

How the learnt information is exploited has a profound effect on the algorithm’s performance. One approach is to directly sample from the model, i.e. generating complete solutions before applying a selection pressure that filters out which complete solutions are better than others \cite{pelikan1999boa}. This effectively conserves correlated variables during future search. A second approach is to use model-informed variation where selection is applied directly after a partial change is made to the solution. This results in an adaptation of the variation operator from substituting single variables to substituting module solutions \cite{watson2011transformations,cox2014solving,mills2014transforming,thierens2010linkage}. DO utilises a model-informed approach as this has been shown to solve optimisation problems that algorithms generating complete solutions from the model cannot \cite{caldwell2017get}.

The application of neural networks to solving optimisation problems has an esteemed history \cite{hopfield1985neural}. Learning heuristics to generalise over a set of CO instances \cite{khalil2017learning,bello2016neural,zhang2000solving} and adapting the learning function to bias future search \cite{hopfield1985neural,boyan2000learning} are popular approaches. DO is different as it uses a DNN to recursively adapt the variation applied. The use of an autoencoder in MBOAs has been attempted \cite{probst2015denoising,churchill2014denoising}, however they limit the autoencoder to a single hidden layer and use the model to generate complete candidate solutions rather than using model-informed variation. DO is the first algorithm to use a deep multi-layered feed-forward neural network to solve CO problems within the framework of MBOAs.

The focus of this paper is to introduce the concept of DO to show how DNNs can be extended to the field of MBOAs. By making this connection we open the opportunity to use the advanced DL tools that have a well-developed conceptual understanding but are not currently applicable to CO and MBOAs. We use two theoretical MAXSAT problems to demonstrate the performance of DO. The Hierarchical Transformation Optimisation Problem (HTOP) contains controllable deep structure that provides clear evaluation of DO during the optimisation process. Additionally, HTOP provides clear evidence that DO is performing as theorised - specifically with regards to the rescaling of the variation operator and the essential requirement for using a layerwise technique. The Parity Modular Constraint optimisation problem (MC\textsubscript{parity}) contains structure with greater than pairwise dependencies and is a simplistic example of how DO can solve problems that current state of the art MBOAs cannot. Finally, DO is used to solve benchmark instances of the Travelling Salesman Problem (TSP) to demonstrate the applicability to CO problems containing characteristics such as non-binary representations and in-feasible solutions. Comparison is made with three heuristic methods in which DOs performance is better.

\section{The Deep Optimisation Algorithm}

\begin{algorithm}
\caption{Deep Optimisation}\label{Alg:DO}
\textbf{Initialise Model}\;
\While{Optimising Model}{ 
 \textbf{Reset Solution}\;
 \While{Optimising Solution}{ 
  Perform model-informed variation to solution\;
  Calculate fitness change to solution\;
  \eIf{Deleterious fitness change}{
    Reject change to solution\;
   }{
   Keep change to solution\;
  }
 }
Update the model using optimised solution as a training example.
}
\end{algorithm}

The Deep Optimisation algorithm is presented in Algorithm \ref{Alg:DO}. The algorithm consists of two optimisation cycles, a solution optimisation cycle and model optimisation cycle, inter-locked in a two-way relationship. The relationship between these two cycles can be understood as a meta-heuristic method where the solution optimiser (heuristic method) is influenced by the model (external control). The solution optimisation cycle is an iterative procedure that produces a locally optimal solution using model-informed variation. The model optimisation cycle is an iterative procedure that updates the connection weights of a neural network to satisfy the learning objective.

DO uses the deep learning Autoencoder model (AE) due to its ability to learn higher-order features from unlabelled data. The encoder and decoder network are updated during training. Only the decoder network is used for generating reconstructions from the hidden layer. DO uses an online learning approach where the learning rate controls the ratio between the exploration and exploitation of the search space.

\subsection{Model Optimisation Cycle}

The AE uses an encoder ($W$) and decoder network ($W'$). The encoder network performs a transformation from the visible units, $X$, to hidden units $H_1$ using a non-linear activation function $f$ on the sum of the weighted inputs: $H_1(X)=f(W_1X + b_1)$, where $W$ and $b$ are the connection weights and bias respectively. The decoder network generates a reconstruction of the visible units, $X_r$, from the hidden units: $X_r(H_1)=f(W_1'H_1 + b_{r1})$ where $W'$ is the transpose of the encoder weights. The backpropagation algorithm is used to train the network using a mean squared error of the reconstruction $X_r$ and input $X$.

A deep AE consist of multiple hidden layers with the encoder performing a transformation from $H_{n-1}$ to $H_n$ defined by $H_n(H_{n-1})=f(W_nH_{n-1} + b_n)$ and the decoder reconstructing $H_{r(n-1)}$ defined by $H_{r(n-1)}(H_{r(n)})=f(W_n'H_{r(n)}+b_{rn})$. DO utilises a layer-wise approach for both training and generating samples. Initially the AE has a single hidden layer and is trained on solutions developed using the naive local search operator. The network then transitions such that the AE consist of two hidden layers and variation is informed by the first layer whilst training updates all connection weights. By constructing a network with less hidden units than visible units creates a regularisation pressure to learn a compression of the training data. At each hidden level, an optimised model will contain a meaningful compression of the lower level relating to higher-orders of organisation. Our experiments show the significance of using a layer-wise approach in comparison to an end-to-end network approach. We employ the notation DO\textsuperscript{n} to differentiate between the number of hidden layers used in the AE.

\subsection{Solution Optimisation Cycle}

The solution optimisation cycle produces locally optimal solutions as guided by model-informed variation. Specifically, a candidate solution $X$ is initialised from a random uniform distribution. A random variation is applied to the candidate solution, forming $X'$, if the variation has caused a beneficial fitness change, or no change to the fitness, the variation is kept, $X=X'$, otherwise the variation is rejected. This procedure is repeated until no further improvements. By repeatedly resetting a candidate solution ensures the training data has sufficiently good coverage of the solution space.

A model-informed variation is generated by performing a bit-substitution to the hidden layer activations at layer $n$, forming $H_n'$. $H_n'$ is then decoded to the solution level using the trained decoder network forming $X'$. The solution optimisation cycle continues as before where the fitness of $X'$ is determined and if there is a fitness benefit, or no change to the fitness, when compared to $X$, $H_n = H_n'$ and $X = X'$, otherwise the change is rejected. A decoded variation made to the hidden layer causes a change to the solution level that exploits the learnt problem structure. Concretely, module-substitutions are constructed by performing bit-substitutions to the hidden layer and decoding to the solution level. At a solution reset, it is important that $H_n$ is an accurate mapping for the current solution state. Therefore the hidden layer $H_n$ is reset using a random distribution $U[-1,1]$. It is then decoded to the solution level $X$ to construct an initial candidate solution. The output of the autoencoder is continuous between the activation values and therefore requires interpreting to a solution. For MAXSAT, DO uses a deterministic interpretation. Specifically if $X'[n] > 0$ then $X'[n] = 1$ else $X'[n] = -1$ where $n$ is the variable index. DO allows neutral changes to the solution. This allows for some degree of drift in the latent space to allow for small effects caused to the decoded output to accumulate and make a meaningful variation to the solution.

As DO uses an unsupervised learning algorithm, for it to learn a meaningful representation the training data must contain information about the hidden problem structure in its natural form. This structure becomes apparent when applying a hill-climbing algorithm to a solution because it ensures that it contains combinations of variables that provide meaningful fitness contributions. Initially, the AE model will have no meaningful knowledge of the problem structure and therefore a model-informed variation is equivalent to a naive search operator. After a transition, DO does not require knowledge of which operator has been initially used, it simply learns and applies its own learnt higher-order variation.

\subsection{Transition: Searching in Combinations of Features}\label{searchfeature}
After the network has learnt a good meaningful representation at hidden layer $n$ the following changes occur to DO, which we term a transition.

\begin{enumerate}
  \item An additional hidden layer $H_{n+1}$ is added to the AE. Previous learnt weights are retained and training updates all weight ($W_1$ to $W_{n+1}$).
  \item The hidden layer used for generating model-informed variation is changed from $H_{n-1}$ to $H_n$. Initialisation of a candidate solution is generated from $H_n$.
\end{enumerate}

Item 1 is analogous to the approach introduced by Hinton and Salakhutdinov \cite{hinton2006reducing} for training DNNs. The layer-wise procedure is important for learning a tractable representation at each hidden layer. The multi-layer network is trained on solutions developed using variation decoded from the layer below the current network depth. This is a significant requirement as DO learns from its own dynamics \cite{watson2011optimization}. There may be many possible mappings in which the problem structure can be represented. Thus deeper layers are not only a representation of higher-order features present in the problem, but are reliant on how the higher-order features have been learnt and exploited, which, in-turn, is determined by the shallower layers. Therefore if the shallower layers do not contain a meaningful representation, then attempting to train or perform variation generated from deeper layers will be ineffective as we prove in our experiments. Item 2 is a layer-wise procedure for generating candidate solutions. The method of generating variation to the solution is the same at any hidden layer. Simply, only the hidden layer where bit-substitutions are performed and decoded from has been changed from $H_{n-1}$ to $H_n$ (to a deeper hidden layer).

This transition procedure is performed recursively until the maximum depth of the autoencoder is reached at which Item 1 is not performed. Like the learning rate, the timing of transition impacts the balance between exploration and exploitation of the search space. Once transitioned not only does the model provide information on how to adapt the applied variation but the solution optimisation cycle provides feedback to the model optimiser. Specifically, correctly learnt features will cause beneficial changes to a solution during optimisation, and therefore will be repeatedly accepted during the solution optimisation cycle and thus repeatedly presented to the model during training, reinforcing the learnt correlations. In contrast, incorrectly learnt features will cause deleterious fitness changes and therefore will not be accepted and thus not present in the training data.

\section{Performance Analysis of Deep Optimisation}
Two theoretical CO problems within the MAXSAT class are specifically designed to demonstrate how DO works and show that DO can solve problems containing high-order dependencies that state of the art MBOA's cannot.

\subsection{How Deep Optimisation Works}\label{sec:howDO}

\begin{table}[]
  \centering
\resizebox{0.6\linewidth}{!}{%

\begin{tabular}{lll}
  \toprule
  \multicolumn{3}{c}{\textbf{HTOP}} \\
  \midrule
  $a$ $b$ $c$ $d$ & $t(a,b,c,d)$ & $f(a,b,c,d)$\\
  \midrule
      1 0 0 0 & 0 0 & 1\\
      0 1 0 0 & 0 1 & 1\\
       0 0 1 0 & 1 0 & 1\\
       0 0 0 1 & 1 1 & 1\\
       Otherwise & - & 0\\
  \bottomrule

\end{tabular}%
}
\caption{HTOP transformation $t$, and fitness function $f$.}
\label{table:HTOP}
\end{table}

The Hierarchical Transformation Optimisation Problem (HTOP) is formed within the MAXSAT class where there objective is to find a solution that satisfies the maximum number of constraints imposed on the problem. HTOP is a consistent constraint problem and has four global optima. HTOP is specifically designed to provide clarity on how DO works with specific regards to the process of rescaling the variation operator to higher-order features and the necessity for a DNN to use a layerwise procedure. HTOP is inspired by Watson's Hierarchical If and only If (HIFF) problem \cite{watson1998modeling} and uses the same recursive construction with an adaptation to cause deep structure. The generalised hierarchical construction is summarised here. The solution state to the problem is
\begin{math}
  x = \{x_1,\ldots,x_i,\ldots,x_N\},
\end{math}
where $x_i \in \{0,1\}$ and $N$ is the size of the problem. $p$ represents the number of levels in the hierarchy and $N_p$ represents the number of building-blocks of length $L_p$ at each hierarchical level. Each block containing $k$ variables is converted into a low-dimensional representation of length $L_p/R$ by a transformation function $t$, where $R$ is the ratio of reduced dimensionality creating a new higher-order string $V^{p+1}=\{V^{p+1}_1,\dots,V^{p+1}_{N_pL_p/R}\}$. In what follows $k=4$ and $R=2$ using the transformation function detailed in Table \ref{table:HTOP}, where a solution to a module is a one-hot bit string.

The transformation function is derived from a machine learning benchmark named the 424 encoder problem \cite{ackley1985learning}. Learning the structure is not trivial and cannot be well approximated by pairwise associations unlike for HIFF. The transformation is applied recursively constructing deep constraint where at each level of hierarchy a one-hot coding is required to be learnt. The null variable is used to ensure that a fitness benefit at the higher level can only be achieved by satisfying all lower level transformations beneath it.

HTOP is pathologically difficult for a bit-substitution hill climber. Satisfying a depth 2 constraint requires coordination of module transformations such that the transformed representations of two modules below construct a one-hot solution, e.g. Module 1 transformation = 01 ($X=0100$) and Module 2 transformation = 00 ($X=1000$). A bit-substitution operation is unable to change a module solution without causing a deleterious fitness change. Therefore a higher-order variation is required that performs substitutions of module solutions. This recursive and hierarchical construction requires the solver to successively rescale the search operator to higher-orders of organisation.

\begin{figure}
	\centering
	\includegraphics[width=0.32\linewidth]{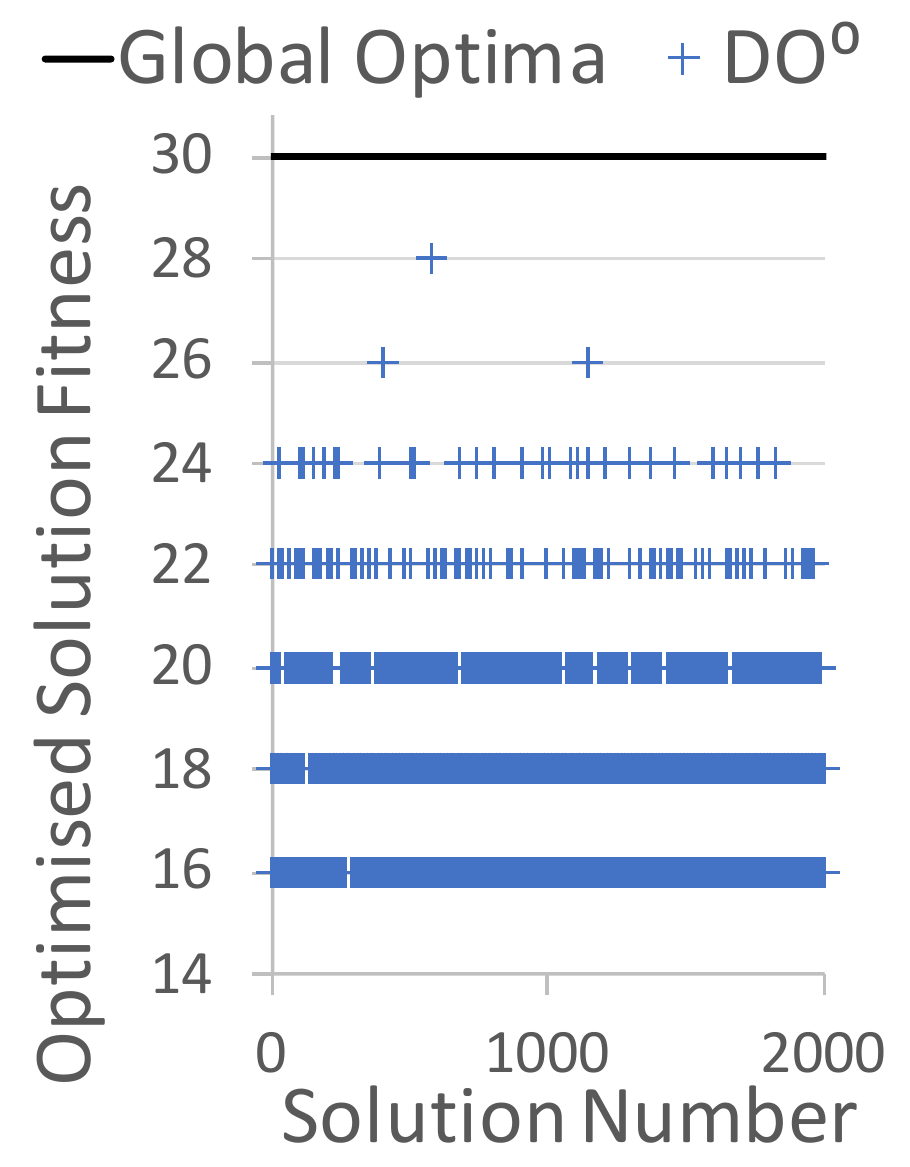}
	\includegraphics[width=0.32\linewidth]{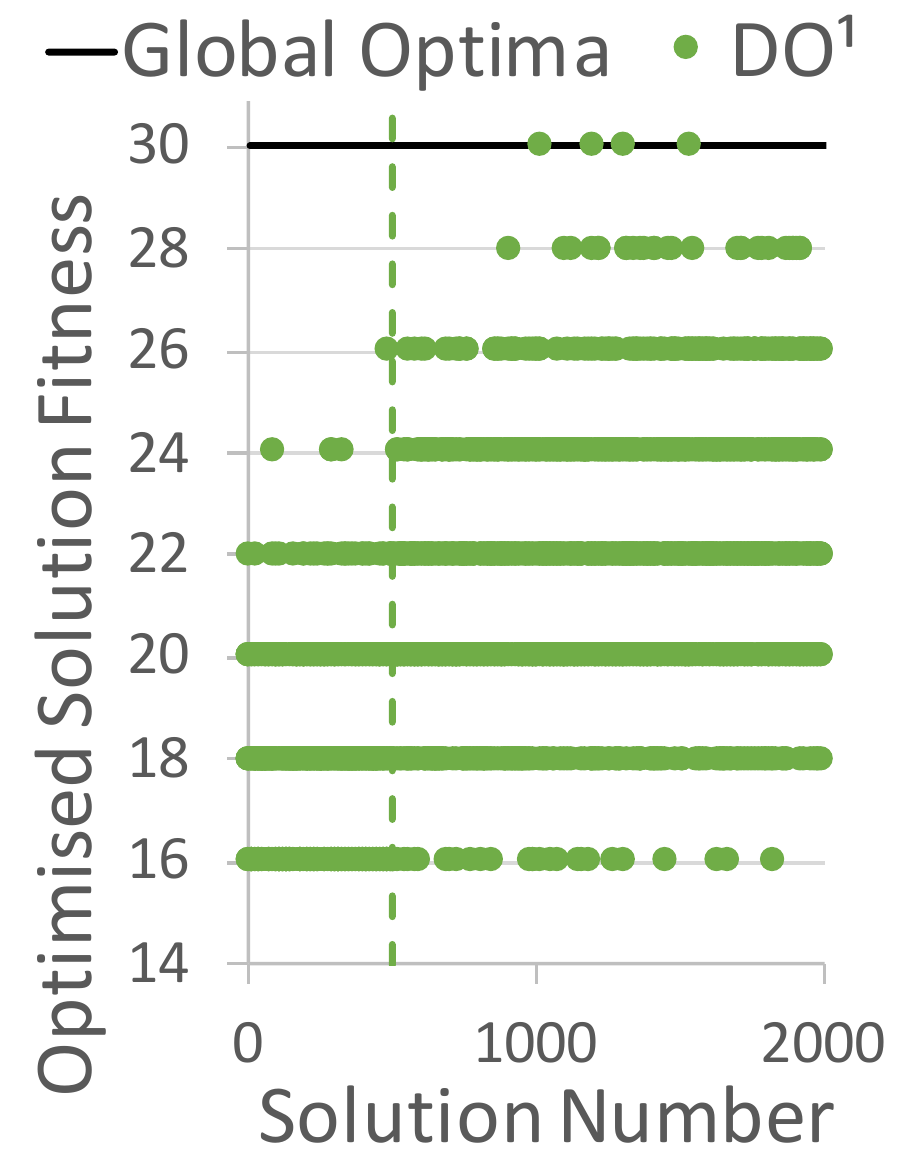}
	\includegraphics[width=0.32\linewidth]{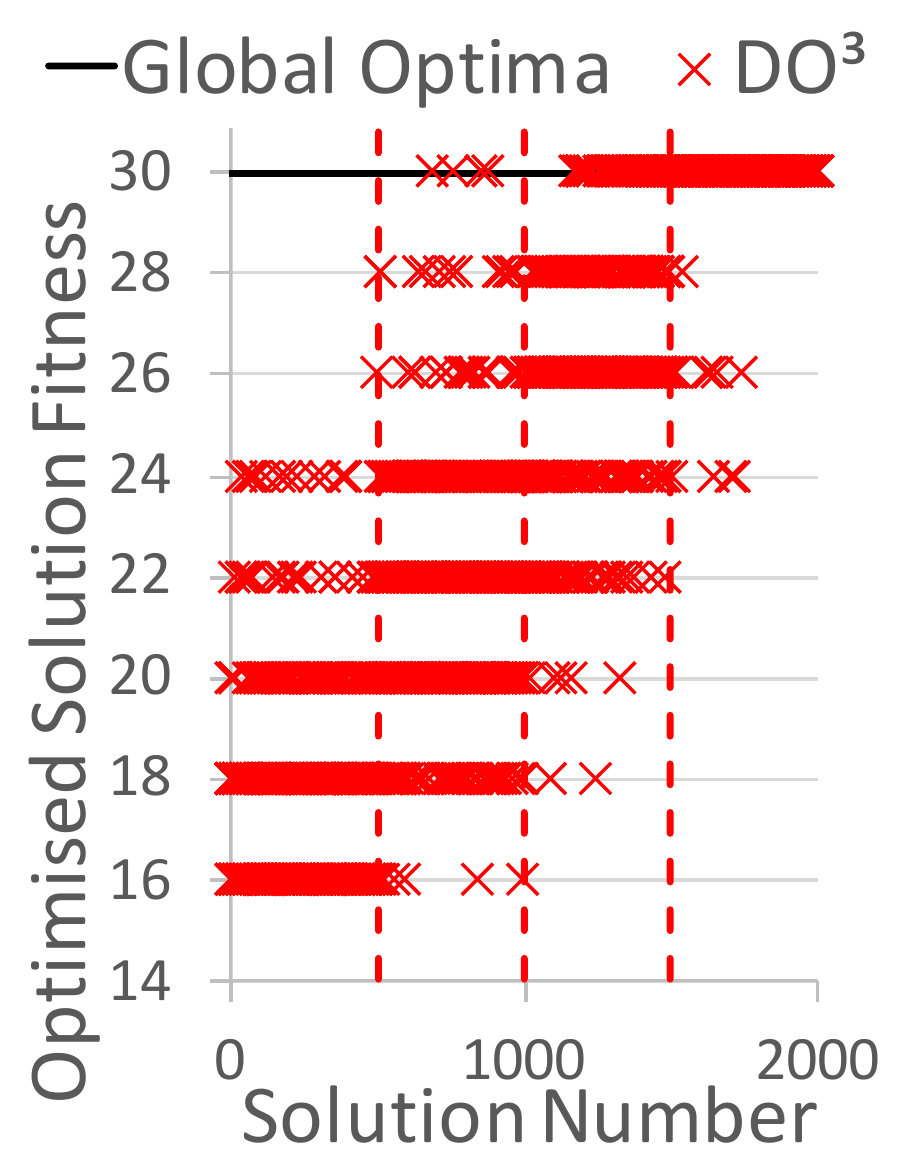}
	\caption{A deep representation allows for learning and exploiting deep structure to find a global optimum.}
	\label{Fig:FitnessHTOP}
\end{figure}

\begin{figure}
	\centering
	\includegraphics[width=\linewidth]{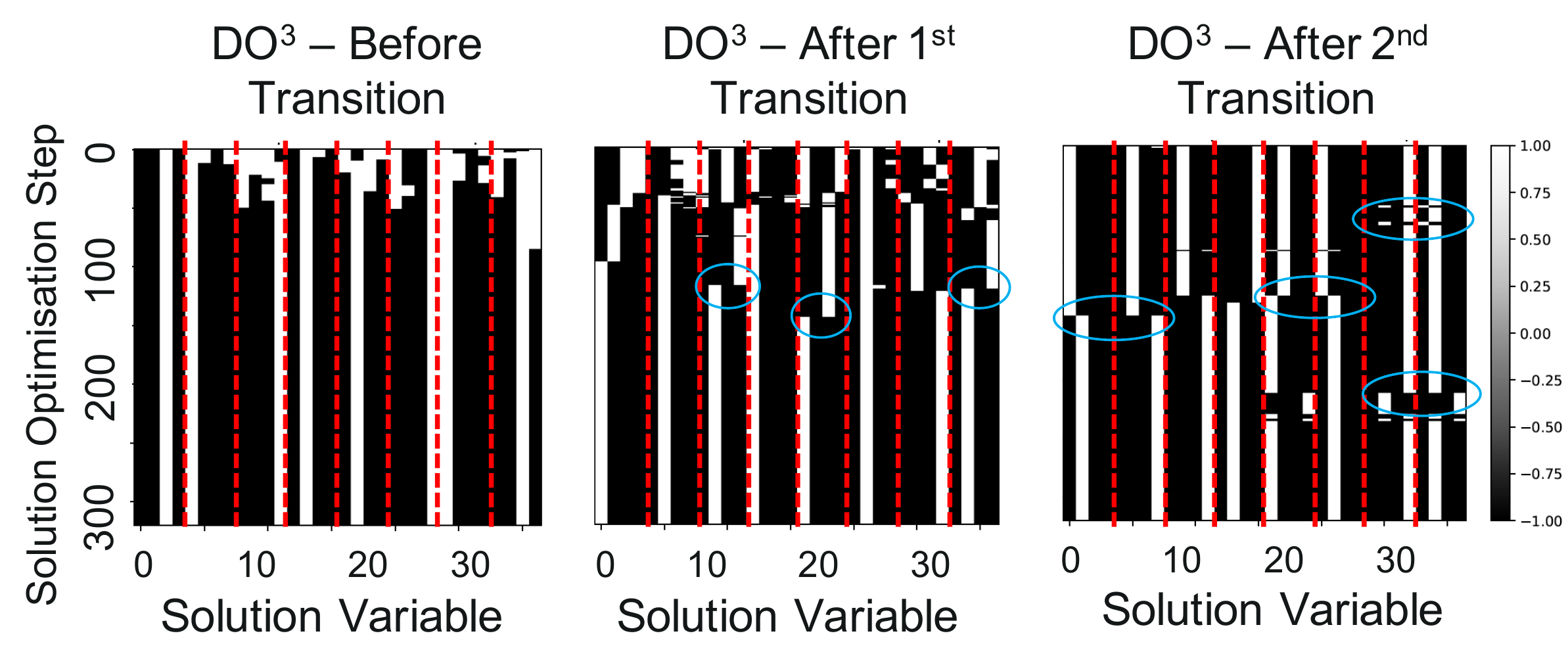}
	\caption{Example solution trajectories during solution optimisation using DO\textsuperscript{3} before transition (left), after 1\textsuperscript{st} transition (middle) and after 2\textsuperscript{nd} transition (right). Variation is adapted from bit-substitutions to module solutions to combinations of module solutions as highlighted by circles. Module boundaries are represented by vertical dashed lines.}
	\label{sFig:VariationHTOP}
\end{figure}

\begin{figure}
\centering
\includegraphics[width=0.32\linewidth]{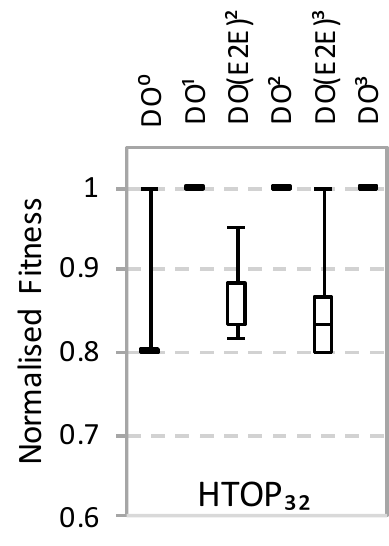}
\includegraphics[width=0.32\linewidth]{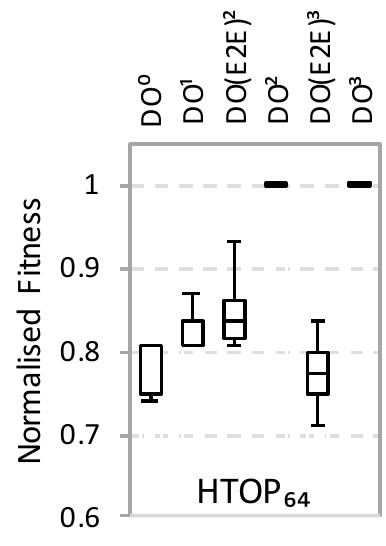}
\includegraphics[width=0.32\linewidth]{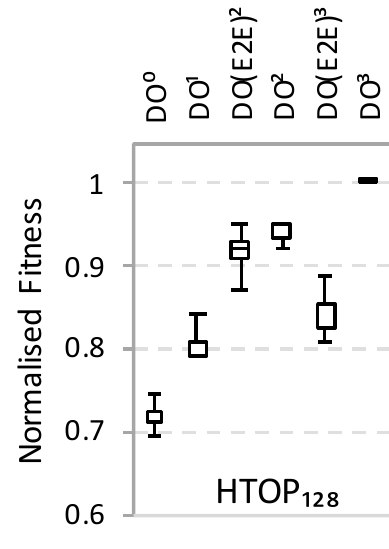}
\caption{Deep representations are consistently better performing than shallow ones, and incremental addition of layers is better than end-to-end learning.}
\label{Fig:shalvdeep}
\end{figure}
A HTOP instance of size 32 (HTOP\textsubscript{32}) is used to demonstrate how DO successfully learns, represents and exploits the deep structure. Further, we show the performance difference between DO\textsuperscript{0} (a bit-substitution restart hill-climbing algorithm), DO\textsuperscript{1} (a restart hill-climber using a shallow network) and DO\textsuperscript{3} (a restart hill-climber using a deep network). The algorithms use 320 steps to optimise a solution and produce a total of 2000 solutions. Figure \ref{Fig:FitnessHTOP} presents the solution fitness after each solution optimisation cycle.

DO\textsuperscript{0} is unable to find a globally optimal solution. It simply gets trapped at local optima as a bit-substitution is insufficient to improve a solution without a deleterious fitness change. DO\textsuperscript{0} therefore has exponential time-complexity to produce a global optima. For DO\textsuperscript{1}, the results show that a single hidden layer is sufficient for finding a global optima. The vertical dashed line illustrates the location of transition. After which DO\textsuperscript{1} is able to perform module substitution without any deleterious fitness effects and thus search for a combination of module solutions to satisfy deeper constraints. However, note that HTOP\textsubscript{32} contains 4 levels of hierarchy and thus a single layer network is not sufficient to fully represent the problem structure. As a result, DO\textsuperscript{1} is unable to perform meta-module substitutions (a change of multiple module solutions simultaneously) and thus the algorithm is unable to converge to a global optima (reliably find a globally optimal solution). DO\textsuperscript{3} shows it is able to find and converge to a globally optimal solution due to having a sufficiently deep network that can correctly learn and represent the full problem structure and thus able to perform meta-module substitutions.

Figure \ref{sFig:VariationHTOP} presents example solution trajectories during the solution optimisation cycle for DO\textsuperscript{3} on HTOP\textsubscript{32}. Initially, before transition, DO only performs a bit-substitution variation and is successful in finding solutions for each module (a one-hot solution per a module), but it is unable to change between module solutions and thus we observe no further changes. After the 1\textsuperscript{st} transition, we observe the variation has been scaled up to allow variation of module solutions (as highlighted by the circles). Now DO can search for the correct combination of modules that satisfy depth 1 constraints without deleterious fitness changes. After the 2\textsuperscript{nd} transition DO is capable of performing meta-module substitutions (module solutions of size 8) enabling it to easily satisfy depth 2 constraints. Hence, we observe, DO is able to learn and represent deep hidden structure and correctly exploit this information in a deep and recursive manner in-order to reduce the dimensionality of the search and adapt the variation operator to solve the problem.

HTOP is a problem that contains deep structure, such that as the size of the problem increases so does the depth of the problem structure. Consequently, a shallow model is unable to solve large instances. Presented in Figure \ref{Fig:shalvdeep} are results showing the fitness for the best solution found, in 10 repeats, by DO using a layer-wised approach: DO\textsuperscript{0}, DO\textsuperscript{1},DO\textsuperscript{2}, DO\textsuperscript{3} (standard method), DO using an end-to-end approach: DO(E2E)\textsuperscript{2}, DO(E2E)\textsuperscript{3}. HTOP\textsubscript{32}, HTOP\textsubscript{64} and HTOP\textsubscript{128} instances are used which have a termination criteria of 800, 2500 and 15000 model optimisation steps respectively.

It is observed that a deeper network is required to solve problems with deeper constraints. Furthermore, the results show the significance of using the DNN in a layer-wise method instead of an end-to-end method. DO(E2E) works by constructing an end-to-end DNN at initialisation and model updates modify all hidden layer connections. A bit-substitution is used to produce the initial training data. At transition, the deepest hidden layer is used for generating a variation. The results clearly show that, whilst the DNN is sufficient to represent the problem structure (as proven by the layer-wise results), using an end-to-end model is not efficient at learning the problem structure so that it can be exploited effectively. Results show that as the DNN gets deeper, using an end-to-end approach produces successively inferior results. A layer-wise approach is therefore essential for DO to work and scale to large problems.

\subsection{Solving what MBOAs Cannot}

\begin{table}[]
\resizebox{\linewidth}{!}{%
\begin{tabular}{|c|c|c|c|c|c|c|}
\hline
\multicolumn{4}{|c|}{Module} & \multicolumn{3}{c|}{Fitness} \\ \hline
1 & 2 & 3 & 4 & \multicolumn{1}{l|}{Within Module} & \multicolumn{1}{l|}{Between Module} & \multicolumn{1}{l|}{Total Fitness} \\ \hline
1000 & 0100 & 1101 & 0000 & 3x1 = 3 & 1 + 1 + 1 = 3 & 3.0003 \\ \hline
1000 & 1000 & 1101 & 1101 & 4x1 = 4 & 2\textsuperscript{2} + 2\textsuperscript{2} = 8 & 4.0008 \\ \hline
1000 & 1000 & 1000 & 1101 & 4x1 = 4 & 3\textsuperscript{2} + 1 =10 & 4.0010 \\ \hline
1000 & 1000 & 1000 & 1000 & 4x1 = 4 & 4\textsuperscript{2} = 16 & 4.0016 \\ \hline
\end{tabular}%
}
\caption{Example solutions to MC\textsubscript{parity}. A global optima is a solution with all modules containing the same parity answer.}
\label{tab:MCparity}
\end{table}

\begin{equation} \label{eq1}
\begin{split}
F & =  \sum_{i=0}^m\begin{cases}1 & \left(\sum_{j=0}^{n}S_j^m\right)\mod 2 = 1\\0 & otherwise\end{cases} \\
 & + p\times\sum_{k=0}^{n/2}\left(\sum_{i=0}^m\begin{cases}1 & S^m == Type_k\\0 & otherwise \end{cases}\right)^2
\end{split}
\end{equation}

The Parity Modular Constraint optimisation problem (MC\textsubscript{parity}) is an adaptation of the Modular Constraint Problem \cite{watson2011optimization} where module solutions are odd parity bit-strings. A problem of size $N$ is divided into $m$ modules each of size $n$. There are $n/2$ sub-solutions per a module and each of the sub-solutions is assigned a type. A fitness point is awarded, for a module, if a module contains an odd parity solution, otherwise no point. A global solution is one where all modules in the problem contain the same parity solution ($n/2$ global optima). The between module fitness is the summation of the squared count of each module solution type present in the whole solution. The fitness function is provided in Equation \ref{eq1} and examples of a solutions fitness is presented in Table \ref{tab:MCparity}. For the scaling analysis performed here we use $n=4$.

Although this problem supports many solutions within each module the smaller fitness benefits of coordinating modules are more rare.
By ensuring the module fitness is much more beneficial than the between module fitness (p $\ll$ 1) requires the algorithm to perform module substitutions of odd parity to follow the fitness gradient to coordinate the module solutions without deleterious fitness effects. If an algorithm cannot learn and exploit the high-order structure of the parity modules then finding a global optima will require exponential time with respect to the number of modules in the problem. Conversely, an algorithm that can will easily follow the fitness gradient to correctly coordinate the module solutions and thus scale polynomial with respect to the number of modules in the problem.

Leading MBOAs such as LTGA, P3 and DSMGA use a dependency structure matrix (DSM) and the mutual information metric to capture variable dependencies. They are successful in capturing module structures containing more than 2 variables however it is hypothesised they are unable to correctly capture structure that contains greater than pairwise dependencies between variables. A simple example being parity. A neural network is capable of learning and capturing higher-order dependencies between variables.

\begin{figure}
\centering
\includegraphics[width=\linewidth]{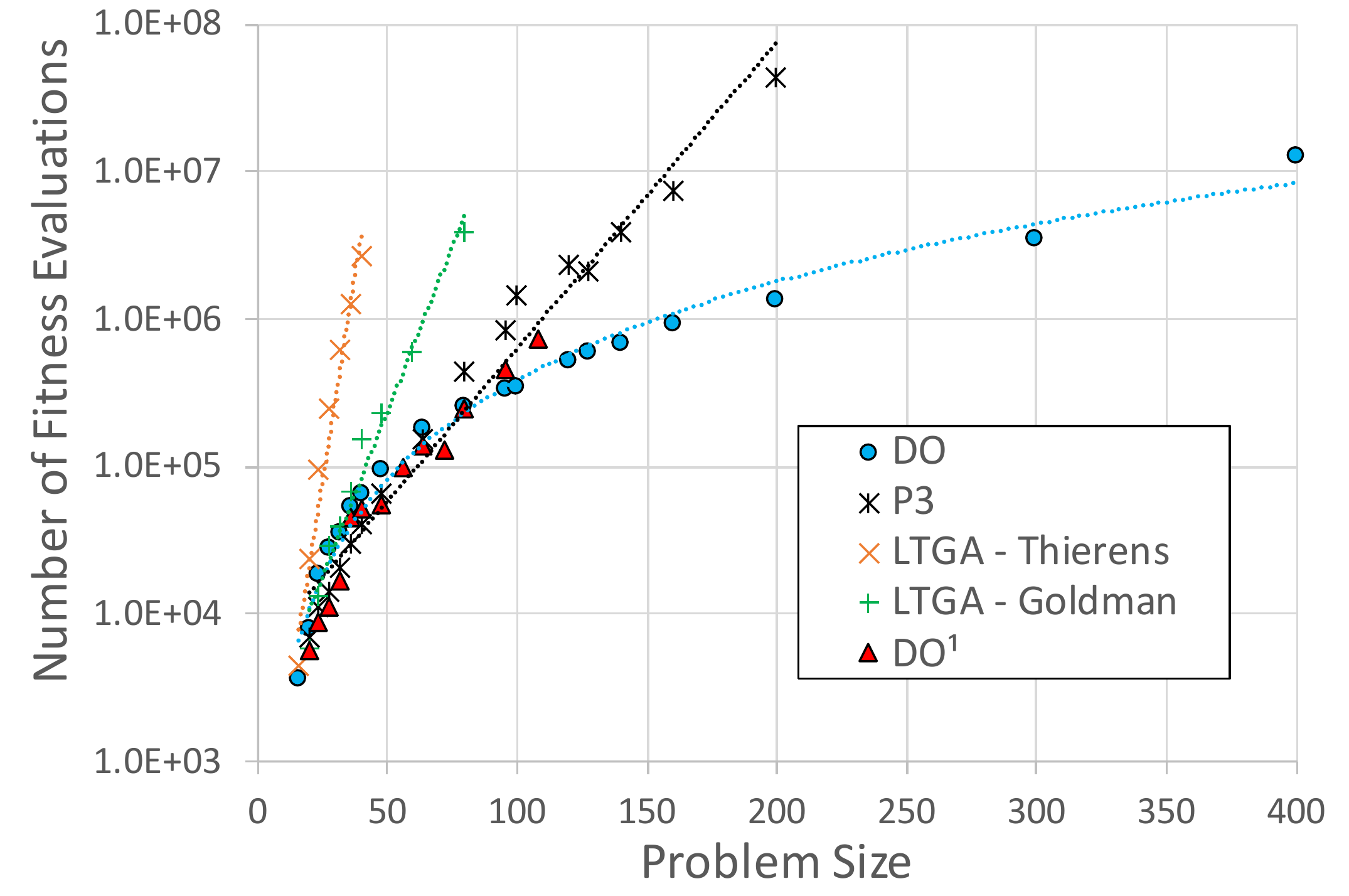}
\caption{DO has polynomial time complexity and MBOAs has exponential time complexity when solving a problem containing high-order dependencies}
\label{Fig:MBOAres}
\end{figure}

For LTGA and DO the parameters are manually adjusted such that all 50 runs produce the global optimum. The results are presented in Figure \ref{Fig:MBOAres}. The data points present the average number of fitness evaluations required to find the global optimum for the 50 independent runs for N $<$ 300 and 10 for N=300 and N=400. For LTGA the population is adjusted manually until a change of 10\% would not cause a failure. For DO, from smallest to largest N, the learning rate varied from 0.05 to 0.0015 and transition from 60 to 1000 solutions. The network topology included up to three hidden layers and used a compression of $\sim 10\%$ at each layer. P3 is parameterless and required no adjustment. Two implementations are included for LTGA, \cite{thierens2010linkage} and \cite{goldman2014parameter}. The differences between both LTGA implementations and P3 are interesting and it is hypothesised to be caused by the way in which solutions can be prioritised according to their fitness for constructing the model. More significantly, LTGA and P3 scale exponentially whereas DO scales polynomial ($\sim N\textsuperscript{2}$).
To verify that the deep structure of DO is necessary a shallow version of DO is included in the results: DO\textsuperscript{1} (limited to a single hidden layer). The scaling appears to be exponential where results could not be achieved for problem instances greater than 108 as the tuning of parameters became extremely sensitive. Whereas with HTOP we can see clearly what the deep structure is that needs to be learnt, in the MC\textsubscript{parity} problem we can see that high-order dependencies defeat other algorithms but are not a problem for DO.

\section{Solving the Travelling Salesman Problem}\label{sec:TSP}

In this section we apply DO to solve the travelling salesman problem (TSP). A solution to a TSP is a route that visits all locations once and returns to the starting location. The optimisation problem is to minimise the total travelling cost. We use 6 TSP instances from the TSP library \cite{reinhelt2014tsplib}: 3 symmetric and 3 asymmetric, and compare with three other heuristic methods. Our aim here is to provide an example of how DO can be successfully used to solve CO problems containing characteristics such as non-binary representations and in-feasible solutions. The results that follow do not show DO outperforming state of the art heuristic methods (these problems are not particularly difficult for Lin-Kernighan Helsgaun algorithm \cite{helsgaun2000effective}) but they do show that DO can find and exploit deep structure that can be used to solve TSP problems better than shallow methods.

\begin{table*}[t]
\centering
\resizebox{0.8\textwidth}{!}{%
\begin{tabular}{|c|c|c|c|c|c|c|c|c|c|c|}
\hline
\multirow{2}{*}{\begin{tabular}[c]{@{}c@{}}Problem \\ Instance\end{tabular}} & \multirow{2}{*}{Type} & \multirow{2}{*}{\begin{tabular}[c]{@{}c@{}}Number \\ Locations\end{tabular}} & \multirow{2}{*}{\begin{tabular}[c]{@{}c@{}}Performance\\DO (\%)\end{tabular}} & \multirow{2}{*}{\begin{tabular}[c]{@{}c@{}}Avg trials \\ req. for DO\end{tabular}} & \multicolumn{3}{l|}{\begin{tabular}[c]{@{}c@{}}Performance using\\same trials as DO (\%)\end{tabular}} & \multicolumn{3}{c|}{\begin{tabular}[c]{@{}c@{}}Performance using\\10000 trials (\%)\end{tabular}} \\ \cline{6-11}
 &  &  & \multicolumn{1}{l|}{} & \multicolumn{1}{l|}{} & \multicolumn{1}{l|}{Swap} & \multicolumn{1}{l|}{Insert} & \multicolumn{1}{l|}{2-Opt} & Swap & Insert & 2-Opt \\ \hline
 fr26 & Sym & 26 & 0 & 30 & 4.6 & 1.2 & 0.2 & 0 & 0 & 0  \\ \hline
 brazil58 & Sym & 58 & 0 & 224 & 17.0 & 4.0 & 0.1 & 0.9 & 0 & 0  \\ \hline
 st70 & Sym & 70 & 0 & 806 & 25.8 & 6.1 & 0.4 & 20.9 & 3.9 & 0.02 \\ \hline
 ftv35 & Asym & 36 & 0 & 112 & 1.6 & 0.3 & 2.7 & 0.8 & 0 & 1.4 \\ \hline
 p43 & Asym & 43 & 0 & 393 & 0.3 & 0.1 & 0 & 0.1 & 0.02 & 0 \\ \hline
 ft70 & Asym & 70 & 0 & 1776 & 17.1 & 4.4 & 26.7 & 7.2 & 2.2 & 23.1 \\ \hline
\end{tabular}%
}
\caption{DO exploits useful structure in TSP problems to find the global optimum (column 4-5) that is not found by a heuristic method within the same number of trials (columns 6-8) nor found easily within 10000 trials (columns 9-11). Values report percentage difference from the global optimum.}
\label{tab:TSPres}
\end{table*}

\subsection{TSP Representation}

To apply DO the TSP solution is transformed into a binary representation using a connection matrix $C$ of size N\textsuperscript{2} where $C\textsubscript{ij}$ represents an edge. $C\textsubscript{ij} = 1$ signifies that $j$ is the next location after $i$ (the remaining entries are 0). There are a total of N connections, where each location is only connected to one other location (not itself) to construct a valid tour. The connection matrix is sparse and we found that normalising the data improved training. This is a non-compact representation but it is sufficient for demonstrating DO's ability for finding and exploiting deep structure.

The output generated by DO is continuous and is interpreted to construct a valid TSP solution. The interpretation is detailed in Algorithm \ref{Alg:intTSP}. There are two stochastic elements included in the routine. The first element is the starting location from which the tour is then constructed. Choosing a random starting position removes the bias associated with starting at the problem defined starting location. The second element is selecting the next location in the tour. The autoencoder is trained such that positive numbers are connections in a tour. A negative output indicates no connection is made between locations. However, for the case when all locations with a positive connection have been used in the tour then, to ensure a feasible solution, the next location is randomly selected from the set of possible locations available. This ensures that learnt sub-tours (building blocks) are correctly conserved during future search and allows the location of the sub-tour to vary within the complete tour (searching in combinations of learnt building blocks). The construction method resembles the method used by Hopfield and Tank \cite{hopfield1985neural} but here we use a max function rather than a probabilistic interpretation.

\begin{algorithm}
\caption{Interpretation for TSP Solution}\label{Alg:intTSP}
Set Tour[0] = select, uniformaly at random, starting position\;
Set ValidLocs = all possible TSP Locations\;
Remove Tour[0] from ValidLocs\;
ConVec = Vector of size N\;
i = 1\;
\Repeat{ValidLocs empty}{
  ConVec = connection vector generated from Autoencoder for location Tour[i-1] \;
  NextLoc = Where(max(ConVec[ValidLocs])) \;
  \eIf{ ConVec[NextLoc] $>$ 0}
	{
		Tour[i] = NextLoc\;
	}{
		Tour[i] = select, uniformaly at random, from ValidLocs\;
	}
	remove Tour[i] from ValidLocs\;
  i++\;}
Cycle tour until Tour[0] = defined start location\;
Tour[i] = defined start location\;
\end{algorithm}

\subsection{Results}

The performance of DO is compared with three local search heuristics: location swap; location insert and  2-opt. The location swap heuristic consists of selecting, at random, two positions in a TSP tour and swapping the locations. The location insert heuristic selects a position in the tour at random, removes the location from the position and inserts it at another random position. The 2-opt heuristic \cite{croes1958method} involves selecting two edge connections between locations, swapping the connections and reversing the sub-tour between the connections. For our experiments, DO used the location insert heuristic before transition as it produces good training data for both symmetric and asymmetric TSP cases. When performing search in the hidden layer local search is also applied.

The results, averaged over 10 runs, are provided in Table \ref{tab:TSPres}. DO solves all TSP instances each time and the number of trials (training examples) are reported in column 5. Columns 6-8 report the percentage difference between the global optimum and the best found solution for a restart hill climber using a heuristic within the trials used by DO to find the global optimum. This demonstrates that DO is exploiting structure as it is able to find the global optimum faster. Note DO used the insert heuristic for all TSP instances, therefore 2-opt can perform better on some small cases as observed. Columns 9-11 report the percentage difference when the heuristic search is allowed 10000 trials. These results further confirm DO is exploiting structure reliably as, especially for the larger instances, the global optimum is not easily found.

\section{Conclusion}
DO is the first algorithm to use a DNN to repeatedly redefine the variation operator used to solve CO problems. The experiments show there exist CO problems that DO can solve that state of the art MBOAs cannot. They also show there exists CO problems that a DNN can solve that a shallow neural network cannot and that using a layer-wise method can solve that an end-to-end method cannot. Further, results show that DO can be successfully applied to CO problems containing characteristics including non-binary representations and in-feasible solutions. This paper thus expands the use of DNN to be applied to CO problems.

DO provides the opportunity to use the advanced deep learning tools that have been utilised throughout the community for other applications of deep learning and are not available to MBOAs, tools such as dropout, regularisation and different network architectures. These tools application have been shown to improve generalisation in conventional DNN tasks and should therefore also improve the ability to learn problem structure and thus DO's ability in solving CO problems. The application in DO remains to be investigated. Whether other network architectures (convolution neural networks, deep belief networks, generative adversarial networks) offer capabilities that are useful for solving some kinds of CO problems, e.g. problems with transposable sub-solutions, is also of interest.

\bibliographystyle{aaai}
\bibliography{DO}

\end{document}